\documentclass{article}

\PassOptionsToPackage{numbers, compress}{natbib}
\usepackage[preprint]{neurips_2026}

\usepackage[utf8]{inputenc} 
\usepackage[T1]{fontenc}    
\usepackage{hyperref}       
\usepackage{url}            
\usepackage{booktabs}       
\usepackage{graphicx}
\usepackage{amsfonts}       
\usepackage{nicefrac}       
\usepackage{microtype}      
\usepackage{xcolor}         

\newcommand{\sreal}{s_\text{real}}
\newcommand{\sfake}{s_\text{fake}}
\newcommand{\preal}{p_\text{real}}
\newcommand{\pfake}{p_\text{fake}}
\newcommand{\ptheta}{\nabla_\theta}
\newcommand{\Rdm}{R_{\text{dm}}}
\newcommand{\Ro}{R_{\text{o}}}
\newcommand{\Adm}{A_{\text{dm}}}

\newcommand{\betadmt}{\beta_{\text{dm},t}}

\usepackage[linesnumbered,ruled,vlined]{algorithm2e}
\SetKwInOut{Input}{Input}
\SetKwInOut{Output}{Output}
\usepackage{amsmath, amssymb, amsfonts}
\usepackage{cleveref}       
\usepackage{xcolor} 
\usepackage[table]{xcolor}
\usepackage{algorithmic} 

\SetCommentSty{mycommfont}
\usepackage{multirow}
\usepackage{array}
\usepackage{pifont}
\usepackage{xcolor}
\definecolor{darkgreen}{RGB}{0,120,0}
\definecolor{darkred}{RGB}{180,0,0}
\newcommand{\cmark}{\textcolor{darkgreen}{\ding{51}}}
\newcommand{\xmark}{\textcolor{darkred}{\ding{55}}}
\newcommand{\oldcmark}{{\ding{51}}}
\usepackage{todonotes}
\usepackage{subcaption}
\usepackage{wrapfig}

\definecolor{commentgreen}{RGB}{0,120,120}
\usepackage{listings}
\usepackage{xcolor}

\title{$\Rdm$: Re-conceptualizing Distribution Matching as a Reward for Diffusion Distillation}

\author{%
\textbf{Linqian Fan\textsuperscript{1,2}\quad Peiqin Sun \textsuperscript{1}\footnotemark[1]\quad Tiancheng Wen \textsuperscript{1}\quad Shun Lu \textsuperscript{1}\quad Chengru Song \textsuperscript{1}} \\[5pt]
\textsuperscript{1}KlingAI Research\quad
\textsuperscript{2}Tsinghua University
}

\begin{document}

\maketitle
\footnotetext[1]{Correspondence to speiqin@gmail.com}

\begin{figure}[h!]
    \centering
        \includegraphics[width=\textwidth]{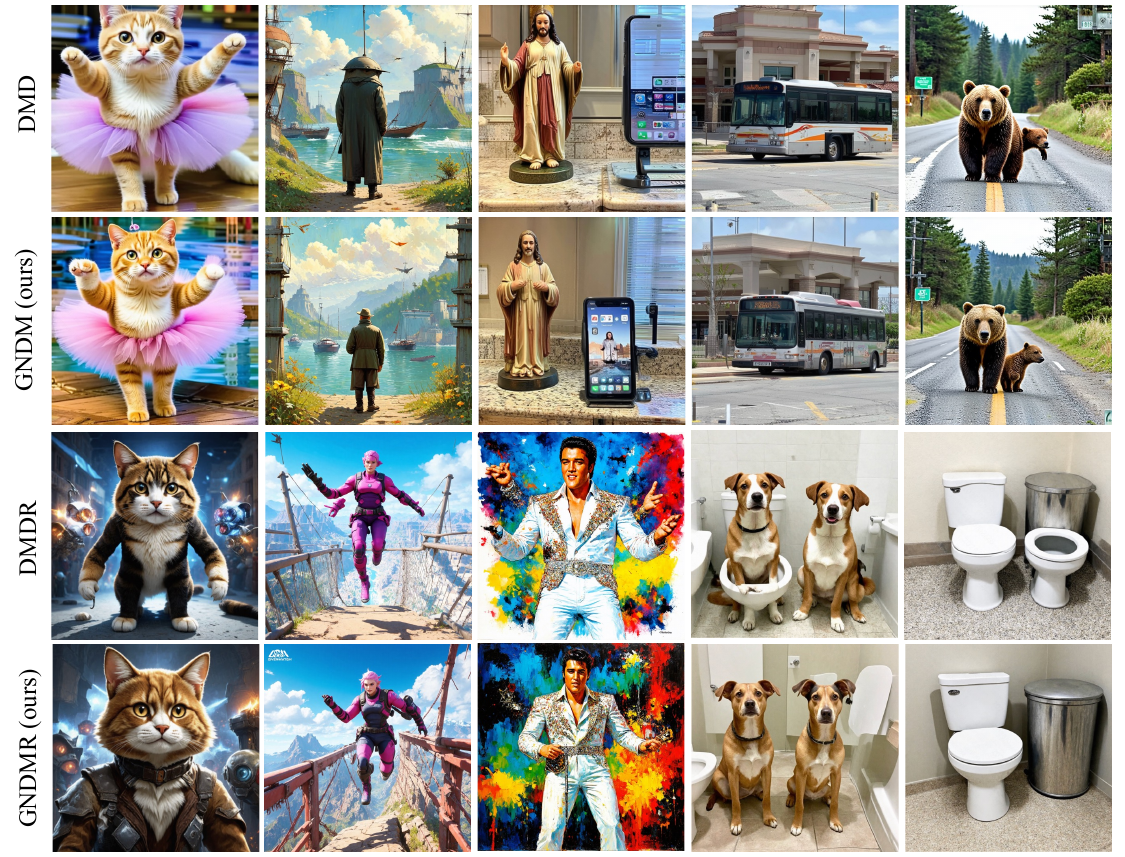}
        \caption{(Top) Samples from 4-step vanilla DMD and our GNDM. (Bottom) Samples from 4-step DMDR and our GNDMR. Our models achieve better perceptual fidelity with fewer artifacts and better details.}
        \label{fig:GN_compare}
\end{figure}

\begin{abstract}
Diffusion models achieve state-of-the-art generative performance but are fundamentally bottlenecked by their slow, iterative sampling process. 
While diffusion distillation techniques enable high-fidelity, few-step generation, traditional objectives often restrict the student’s performance by anchoring it solely to the teacher. Recent approaches have attempted to break this ceiling by integrating Reinforcement Learning (RL), typically through a simple summation of distillation and RL objectives.
In this work, we propose a novel paradigm by \textbf{re-conceptualizing distribution matching as a reward}, denoted as $\Rdm$. This unified perspective bridges the algorithmic gap between Diffusion Matching Distillation (DMD) and RL, providing several primary benefits: \textbf{(1) Enhanced Optimization Stability:} We introduce Group Normalized Distribution Matching (GNDM), which adapts standard RL group normalization to stabilize $\Rdm$ estimation. By leveraging group-mean statistics, GNDM establishes a more robust and effective optimization direction. \textbf{(2) Seamless Reward Integration:} Our reward-centric formulation inherently supports adaptive weighting mechanisms, allowing for the fluid combination of DMD with external reward models. \textbf{(3) Improved Sampling Efficiency:} By aligning with RL principles, the framework readily incorporates Importance Sampling (IS), leading to a significant boost in sampling efficiency.
Extensive experiments demonstrate that GNDM outperforms vanilla DMD, reducing the FID by 1.87. Furthermore, our multi-reward variant, GNDMR, surpasses existing baselines by striking an optimal balance between aesthetic quality and fidelity, achieving a peak HPS of 30.37 and a low FID-SD of 12.21. Ultimately, $\Rdm$ provides a flexible, stable, and efficient framework for real-time, high-fidelity synthesis. Codes are coming soon.
\end{abstract}

\section{Introduction}
\label{sec:intro}
Diffusion models\cite{sd3,DiT,ldm,ddpm,score-based} have established a new state-of-the-art in generative modeling, but they are fundamentally limited by their iterative sampling process, which incurs significant computational overhead. To achieve high-fidelity, real-time synthesis, researchers have explored various distillation strategies\cite{guided_distill, progress_distill, improved_cm, sim, guided_sid,dmd,dmd2}. Among these, Distribution Matching Distillation (DMD)\cite{dmd,dmd2} has been widely adopted due to its exceptional ability to enable high-fidelity generation in just one or a few steps.

\begin{figure}[htp]
    \centering
        \includegraphics[width=\textwidth]{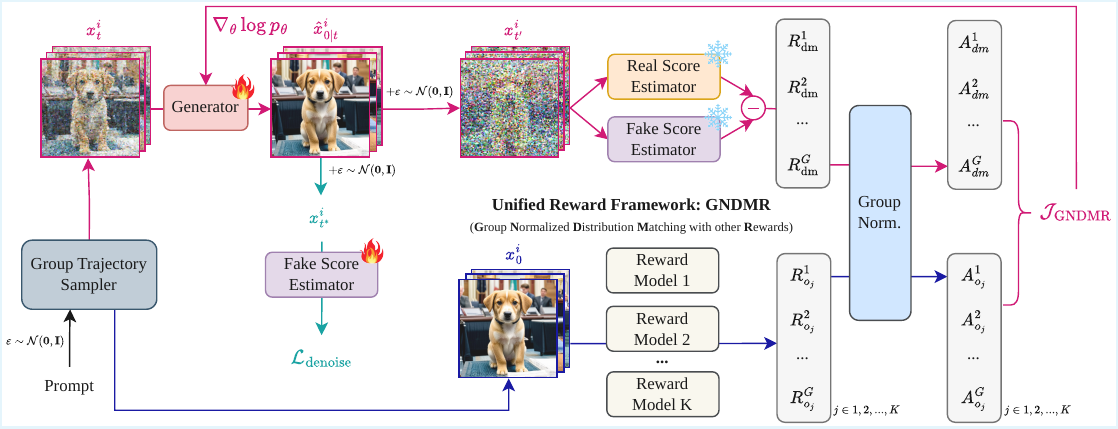}
        \caption{Our unified reward framework GNDMR. After re-conceptualizing distribution matching as a reward, $\Rdm$ and other rewards perform GRPO simultaneously.}
        \label{fig:pipeline}
\end{figure}

However, traditional distillation objectives inherently bottleneck the performance of student models, as the optimization target is derived exclusively from a pretrained teacher \cite{dmdr,adm}. To break this performance ceiling, recent studies \cite{lasro,dipp,dmdr} have integrated Reinforcement Learning (RL) with diffusion distillation. However, these approaches typically rely on \textbf{a naive linear combination} of RL objectives and distillation losses. Departing from these paradigms, we adopt a fundamentally different perspective: \textbf{re-conceptualizing distribution matching as a reward}. This formulation allows the distillation objective to be seamlessly integrated and jointly optimized with task-specific rewards within a unified reward framework, as illustrated in \Cref{fig:pipeline}.
Building upon this conceptual shift, we formally define distribution matching as $\Rdm$. This transition reveals a critical challenge: as image noise increases, the variance of $\Rdm$ amplifies significantly. This escalating variance destabilizes the estimated optimization direction and leads to inefficient training, a phenomenon we analyze in \Cref{sec:revisit_rdm}. To mitigate this, we draw inspiration from established RL practices and introduce Group Normalization (GN) to provide a stabilized, superior optimization gradient, resulting in Group Normalized Distribution Matching (GNDM). Moreover, framing DMD strictly as a reward maximization problem unlocks several critical advantages when integrating with other rewards: First, $\Rdm$ naturally informs the design of an effective adaptive weighting function to balance multiple objectives. Second, it seamlessly incorporates Importance Sampling (IS), which significantly improves sampling efficiency during training. Most importantly, this paradigm shift effectively constructs a algorithmic bridge between diffusion distillation and the broader RL ecosystem. We can now effortlessly leverage established RL techniques\cite{ppo,flow-grpo,adrpo,gdpo} to further refine the distillation process under a single, unified mathematical umbrella.

Our contributions are summarized as follows:
\begin{itemize}
    \item We propose $R_{\text{dm}}$, which natively incorporates powerful RL techniques into the diffusion distillation pipeline. This unification resolves the optimization conflicts inherent in prior joint-training methods and enables more intuitive control over training dynamics.
    \item We introduce Group Normalized Distribution Matching (GNDM) to provide high-fidelity directional guidance and propose a unified reward framework (GNDMR) to holisticlly optimize the distillation process.
    \item We conduct extensive experiments demonstrating that GNDM achieves superior distillation performance over vanilla DMD. Furthermore, our unified GNDMR framework surpasses existing baselines, yielding a highly optimal balance between visual aesthetics and distillation efficiency.
\end{itemize}

\section{Related Work}
\textbf{Distribution Matching Distillation.} Distribution Matching Distillation (DMD)\cite{dmd}  is a foundation work that applies score-based distillation to large-scale diffusion models. A lot of follow-up work emerged to enhance its stability, theoretical grounding, and generation quality. DMD2\cite{dmd2} integrates a GAN loss to eliminate the reliance on costly paired regression data. Flash-DMD\cite{flash-dmd} designs a timestep-aware strategy and incorporates pixel-GAN to achieve faster convergence and stable distillation. While numerous works need GAN to achieve better performance, TDM\cite{tdm} combines trajectory distillation and distribution matching for better alignment and eliminates GAN. Decoupled DMD\cite{decoupled-dmd} mathematically decomposes the DMD objective, revealing that Classifier-Free Guidance (CFG) augmentation acts as the primary generative engine while distribution matching serves as a regularizer, enabling optimized decoupled noise schedules. DMDR\cite{dmdr} combines DMD with RL, utilizing the distribution matching loss as a regularization mechanism to safely allow the student generator to explore and ultimately outperform the teacher.

\noindent\textbf{Reinforcement Learning for Diffusion Models.} Reinforcement learning (RL) has been widely adopted to align diffusion models with human preferences. Various algorithms have been developed for this purpose: ReFL\cite{refl} achieves strong performance but inherently relies on differentiable reward models, whereas Direct Preference Optimization (DPO)\cite{dpo} optimizes the policy using pairwise data. Alternatively, Denoising Diffusion Policy Optimization (DDPO)\cite{ddpo} requires only a scalar reward, a process that Group Relative Policy Optimization (GRPO)\cite{flow-grpo} further simplifies by eliminating the critic model via group normalization. Applying RL to distilled diffusion models is typically treated as an independent post-training phase: Pairwise Sample Optimization (PSO)\cite{pso} first adapted policy optimization for distilled models, and Hyper-SD\cite{hyper-sd} incorporated human feedback to further boost accelerated generation. Breaking away from decoupled training, DMDR\cite{dmdr} introduced the first framework to simultaneously optimize both DMD and RL objectives. Our proposed method is also built upon the DMDR\footnote{We discuss only the GRPO-based variant of DMDR.} framework.

\section{Preliminaries}
\subsection{Distribution Matching Distillation}
The goal of Distribution Matching Distillation (DMD)\cite{dmd} is to distill a multi-step diffusion model (teacher) into a high-fidelity, few-step generator (student) $G_\theta$. The primary objective of DMD is Distribution Matching Loss (DML), which minimizes the reverse-KL divergence between the teacher's distribution $\preal$ and student's distribution $\pfake$. The gradient of DML is:
\begin{align}
\label{eq:DMD}
\nabla_\theta \mathcal{L}_{\text{DMD}} =-\mathbb{E}_{\varepsilon,t'} \left( s_{\text{real}}(x_{t'}) - s_{\text{fake}}(x_{t'}) \right)\nabla_\theta G_\theta(\varepsilon) ,
\end{align}
where $\varepsilon \sim \mathcal{N}(\mathbf{0}, \mathbf{I})$, $t' \sim \mathcal{U}(T_{\text{min}}, T_{\text{max}})$ and $x_t'$ is the diffused sample obtained by injecting noise into $x_0=G_\theta(\varepsilon)$ at diffused time step $t'$. $s_\text{real}$ and $s_\text{fake}$ are score functions given by score estimator $\mu_\text{real}$ and $\mu_\text{fake}$. During training, $\mu_\text{fake}$ with parameter $\psi$ is initialized with $\mu_\text{real}$, and updating to track the distribution of $G_\theta$ through denoising diffusion objective:
\begin{equation}\label{eq:denoise}
    \mathcal{L}_{\text{denoise}} = ||\mu_\text{fake}^\psi(x_{t'},t')-x_0||^2_2.
\end{equation}
In few-step generation training, $G_\theta(\varepsilon)$ is revised by backward simulation as introduced from DMD2\cite{dmd2}. We follow SDE-based inference methods, starting from Standard Gaussian noise, it iteratively perform denoising $\hat x_{0\vert t} = G_\theta(x_t)$ and noising $x_{t-1} = \alpha_{t-1} \hat x_{0\vert t} + \sigma_{t-1} \varepsilon$. Thus, we have:
\begin{equation}
\label{eq:pt}
    p_\theta(x_{t-1}|x_t) = \mathcal{N}(\alpha_{t-1} G_\theta(x_t), \sigma_{t-1}^2\mathbf{I}),
\end{equation}
and we can redefine the gradient of one-step DML in \Cref{eq:DMD} to multi-step DML:
\begin{equation}
\label{eq:DMD2}
\nabla_\theta \mathcal{L}_{\text{DMD}} = -\mathbb{E}_{t',x_t\sim G_\theta} \left( s_{\text{real}}(x_{t'}) - s_{\text{fake}}(x_{t'}) \right)\nabla_\theta G_\theta(x_t)
\end{equation}
\subsection{Denoising Diffusion Policy Optimization}
\label{sec:ddpo}
Following Denoising Diffusion Policy Optimization (DDPO)\cite{ddpo}, we map the denoising process to the following multi-step Markov decision process (MDP):
\begin{align*}
    &s_t \triangleq (t,x_t,c),\quad \pi(a_t \mid s_t) = p_\theta(x_{t-1} \mid x_t,c), \quad P(s_{t+1} \mid s_t, a_t) \triangleq \bigl(\delta_c,\delta_{t-1}, \delta_{x_{t-1}}\bigr),\\
    &a_t \triangleq x_{t-1},\quad  \rho_0(s_0) \triangleq \bigl(c,\delta_T, \mathcal{N}(\mathbf{0}, \mathbf{I})\bigr), \quad R(s_t, a_t) \triangleq 
    \begin{cases}
        r(x_0, c), & t = 0, \\
        0, & \text{otherwise}.
    \end{cases}\\
\end{align*}
Where $\delta_x$ is Dirac delta distribution and $T$ denoted the length of sampling trajectories.

After collecting denoising trajectories $\{x_T,x_{T-1},...,x_0\}$ and likelihoods $\log p_\theta$, we can use policy gradient estimator depicted in REINFORCE\cite{reinforcement,monte} to update parameter $\theta$ via gradient descent:
\begin{equation}
\label{eq:DDPO}
    \nabla_\theta \mathcal{J}_\text{DDPO}
    =
    \mathbb{E}\left[ r(x_0)
        \sum_{t=1}^{T}
        \nabla_\theta \log p_\theta(x_{t-1} \mid x_t)\right],
\end{equation}
To overcome the limitation where optimization is confined to one step per sampling round due to the on-policy requirement of the gradient, we utilize an importance sampling estimator\cite{approximately}. This formulation facilitates multi-step optimization by reweighting gradients from trajectories produced by $\theta_{\text{old}}$:
\begin{equation}
\nabla_{\theta} \mathcal{J}_{\text{DDPO}_\text{IS}} = \mathbb{E} \left[ r(x_0)\sum_{t=1}^{T} \frac{p_{\theta}(x_{t-1} \mid x_t)}{p_{\theta_{\text{old}}}(x_{t-1} \mid x_t)} \nabla_{\theta} \log p_{\theta}(x_{t-1} \mid x_t)  \right]
\end{equation}
In this context, the expectation is taken with respect to the denoising sequences generated under the previous parameter set $\theta_{\text{old}}$.

Note that for ease of observation we omit the condition $c$ in all formulas. \Cref{eq:DDPO} is similar in form to \Cref{eq:DMD2}, which inspires us to regard term $(\sreal-\sfake)$ as a reward.

\section{Methodology}

\subsection{$\Rdm$: Distribution Matching as a Reward}

The key to establishing the connection between \Cref{eq:DMD2} and \Cref{eq:DDPO} is uncovering the relationship between $\ptheta G_\theta(x_t)$ and $\ptheta \log p_\theta(x_{t-1}|x_t)$, which are intrinsically linked through \Cref{eq:pt}. By taking the log-derivative of \Cref{eq:pt}, we obtain the following identity:
\begin{equation}
\ptheta \log p(x_{t-1} \mid x_t) = \frac{x_{t-1} - \mu_\theta(x_t)}{\sigma_{t-1}^2} \cdot \ptheta \mu_\theta(x_t)
\end{equation}
where $\mu_\theta(x_t)=\alpha_{t-1}G_\theta(x_t)$. Consequently, if we set 
\begin{equation}
\label{eq:Rdm}
\Rdm(x_t,x_{t-1},t') = \frac{\sreal(x_{t'})-\sfake(x_{t'})}{x_{t-1} - \mu_\theta(x_t)}\cdot \frac{\sigma_{t-1}^2}{\alpha_{t-1}},
\end{equation}
the relationship between the $\ptheta G_\theta(x_t)$ and $\ptheta \log p_\theta(x_{t-1}|x_t)$ can be formulated as:
\begin{equation*}
\Rdm(x_t,x_{t-1},t')\ptheta \log p_\theta(x_{t-1}|x_t)=(\sreal(x_{t'})-\sfake(x_{t'})) \nabla_\theta G_\theta(x_t).
\end{equation*}
We define the policy gradient objective function for the distillation-based policy as:
\begin{equation}
\label{eq:ddpo_dm}
\nabla_{\theta} \mathcal J_{\text{DDPO}_{\text{DM}}} = \mathbb{E}_{t',x_t\sim G_\theta}\left[ \Rdm(x_t,x_{t-1},t') \nabla_{\theta} \log p(x_{t-1}|x_t)\right],
\end{equation}
Comparing to \Cref{eq:DDPO}, we use only one sample from the trajectory instead of all samples, ensuring consistency with the traditional DML as in \Cref{eq:DMD2}. Finally, We have $\nabla_{\theta} \mathcal J_{\text{DDPO}_{\text{DM}}} = -\ptheta \mathcal L_{DMD}$ strictly established.

\subsection{Revisiting the Distribution Matching Reward}\label{sec:revisit_rdm}
Next, we further explore the meaning of our new-defined distribution matching reward $\Rdm$. By substituting $x_{t-1} - \mu_\theta(x_t) = \sigma_{t-1} \varepsilon^x$ ($\varepsilon^x$ is the standard gaussian noise which sampled to obtain $x_{t-1}$), we can rewrite \Cref{eq:Rdm} as 
\begin{equation}\label{eq:Rdm2}
    \Rdm(t,t')=R_s(t')\cdot \frac{\sigma_{t-1}}{\alpha_{t-1}\varepsilon^x},
\end{equation}
where $R_s(t') := \sreal(x_{t'})-\sfake(x_{t'})$. 

\begin{wrapfigure}{r}{0.3\linewidth}
    \includegraphics[width=\linewidth]{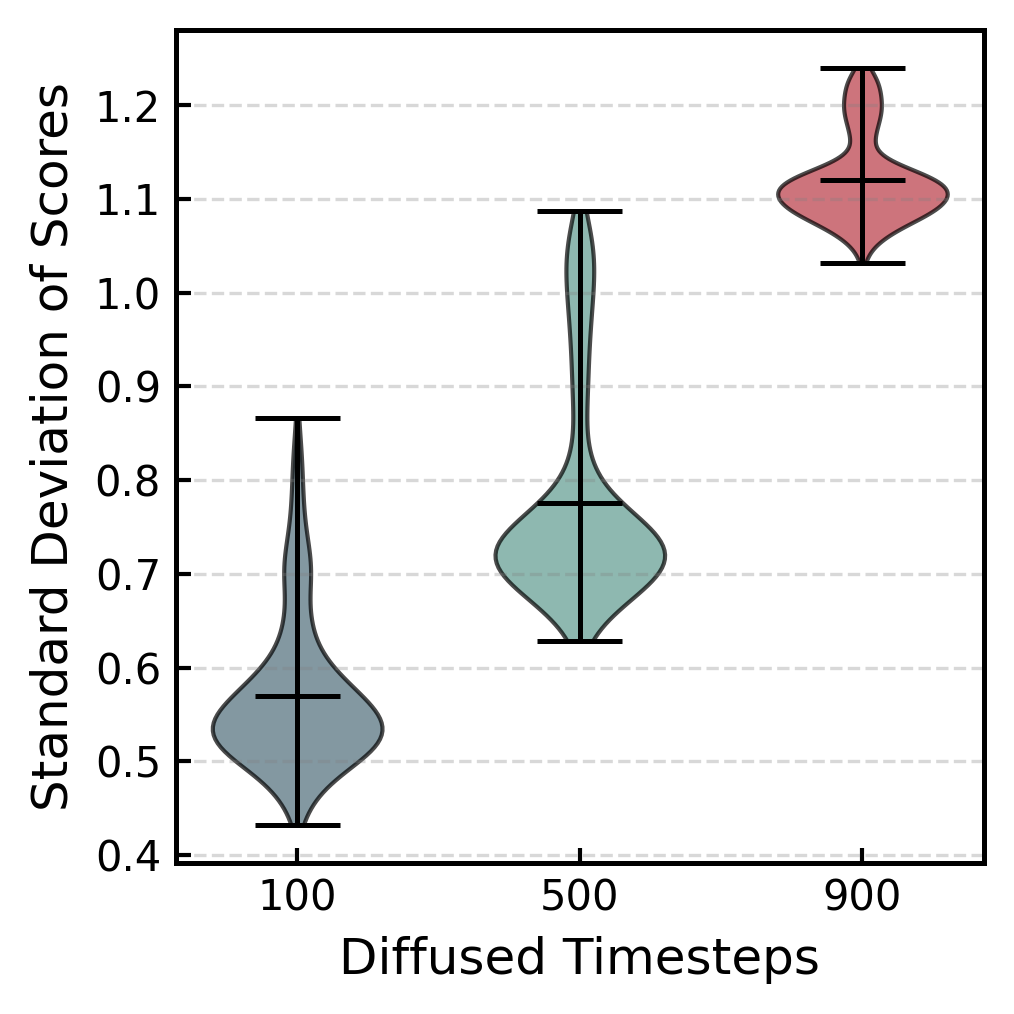}
    \caption{Larger diffused timesteps, higher score variance.}
    \vspace{-15pt}
    \label{fig:std}
\end{wrapfigure}

The vector $R_s(t')$ functions as the critical guidance term that drives the generation of $G_\theta$ toward the manifold of $p_{\text{real}}$ and away from $p_{\text{fake}}$. 
However, $R_s(t')$ fluctuates across diffused samples $x_{t'}$ at timesteps $t'$. Specifically, as $t'$ increases, the diffused sample $x_{t'}$ becomes increasingly blurred, resulting in more ambiguous directional information provided by $R_s(t')$, as shown in \Cref{fig:std}. Although recent works leverage the sample weighting mechanisms introduced by DMD\cite{dmd} to enhance stability and fidelity, the variance of $R_s(t')$ is higher at larger values of $t'$, which can mislead the optimization trajectory, thereby increasing the difficulty of effective model distillation. Further analysis in \Cref{sec:abl_interval}.

Beyond the optimization challenges posed by the high variance at larger timesteps, the fundamental objective of the distribution matching reward $\Rdm$ diverges significantly from standard RL paradigms. Unlike conventional reward metrics such as HPS\cite{hps} or CLIP Score\cite{clipscore}, where unbounded maximization often exacerbates reward hacking, $\Rdm$ essentially acts as a divergence measure that is optimized to approach zero. This stabilization dynamic is conceptually analogous to PREF-GRPO\cite{pref-grpo}, where win-rate converges to 0.5. This fundamental shift from absolute maximization to distribution alignment introduces a critical advantage: $\Rdm$ provides an inherent safeguard against over-optimization, and it is a well-defined reward for regularization. 

\subsection{Group Normalized Distribution Matching}
After re-concepting the distribution matching as a reward, we aim to address the inaccurate estimation issue in the calculation of $R_s(t')$. As Group Normalization (GN) on $\Rdm$ allows for a more stable estimation of the reward direction by the mean-subtraction mechanism, we naturally extend $\Rdm$ to a GRPO setting.

Recall the MDP formulation defined in \Cref{sec:ddpo}. The generator $G_\theta$ samples a groups of G individual images $\{x_0^i\}_{i=1}^G$ and the corresponding trajectories $\{(x_T^i,x_{T-1}^i,...,x_0^i)\}_{i=1}^G$. The advantage of $i$-th sample at trajectory step $t$ and diffused timestep $t'$ is 
\begin{equation}
\label{eq:Adm}
    A_{\text{dm},t}^{i,t'} = \frac{\Rdm(x_t^i,x_{t-1}^i,t')-\text{mean}(\{\Rdm(x_t^i,x_{t-1}^i,t')\}_{i=1}^G)}{\text{std}(\{\Rdm(x_t^i,x_{t-1}^i,t')\}_{i=1}^G)}.
\end{equation}
The policy is updated by maximizing the Group Normalized Distribution Matching (GNDM) objective:
\begin{equation}
\mathcal{J}_{\text{GNDM}}(\theta) = \mathbb{E}_{t,t',{\{x^i\}_{i=1}^G} \sim G_{\theta_\text{old}}} \left[ f(r, A_\text{dm}, \theta, \eta) \right],
\label{eq:grpo_dm}
\end{equation}
where
\begin{align}
f(r, A_\text{dm}, \theta, \eta) &= \frac{1}{G} \sum_{i=1}^G
\min \left( r_t^i(\theta) A_{\text{dm},t}^{i,t'}, \text{clip}\bigl(r_t^i(\theta), 1-\eta, 1+\eta\bigr) A_{\text{dm},t}^{i,t'} \right),\\
r_t^i(\theta) &= \frac{p_\theta(x_t^i \mid x_{t-1}^i)}{p_{\theta_{\text{old}}}(x_t^i \mid x_{t-1}^i)}. \notag
\end{align}
In each group, we share the generator timestep $t$ and diffused timestep $t'$, separately. It maximizes component diversity while ensuring consistency within the group. We prove it is effective in ablation study \Cref{sec:abl_timestep}.
Furthermore, we introduced importance sampling in \Cref{eq:grpo_dm}, which enables to increase sampling efficiency while maintaining performance by update generator multiple times in once sampling, as discuss in \Cref{sec:IS}.
\subsection{GNDM with Other Rewards}
To further improve the generative quality and alleviate mode seeking, following DMDR\cite{dmdr}, we introduce other reward $\Ro$ in addition to the $\Rdm$. As the same to \Cref{eq:Adm}, We employ GN to $\Ro$:
\begin{equation}
\label{eq:Ao}
    A_{\text{o},t}^i = \frac{\Ro(x_0^i)-\text{mean}(\{\Ro(x_0^i)\}_{i=1}^G)}{\text{std}(\{\Ro(x_0^i)\}_{i=1}^G)}.
\end{equation}
Unlike $A_{\text{dm},t}^{i,t'}$, which is calculated as a dense reward on latent trajectory, $A_{\text{o},t}^i$ is calculated as a sparse reward on final image.

The total advantage of $i$-th sample $A_{\text{sum},t}^i$ at generation time $t$ can be defined as the combination of the DM-derived advantage $A_{\text{dm},t}^{i,t'}$ and $K$ auxiliary advantages $A_{\text{o}_j,t}^i$:
\begin{equation}\label{eq:Asum}
    A_{\text{sum},t}^i = A_{\text{dm},t}^{i,t'} + \sum_{j=1}^K w_jA_{\text{o}_j,t}^i,
\end{equation}
where $w_j$ is the weighting funciton for $j$-th auxiliary reward. The final GRPO objective with multi-rewards is:
\begin{equation}\label{eq:loss_GNDMR}
    \mathcal{J}_{\text{GNDMR}}(\theta) = \mathbb{E}_{t,t',{\{x^i\}_{i=1}^G} \sim G_{\theta_\text{old}}} \left[ f(r, A_\text{sum}, \theta, \eta) \right].
\end{equation}

\subsection{Adaptive Weight Design in Practice Implement}
In practice, the term $x_{t-1} - \mu_\theta(x_t)$ introduces significant stochasticity and potential numerical instability, as $x_{t-1}$ is obtained by adding gaussian noise from $\mu_\theta(x_t)$. To address this, we define a stabilized reward term $R_{\text{dm}}$ by applying a sign-based normalization to the denominator to guarantee positive correlation. We also consider the weighting function proposed by DMD\cite{dmd}, as we found it can improve distillation efficiency. The final distillation matching reward can be defined as
\begin{equation}\label{eq:Rdm_practice}
R_{\text{dm}}(x_t,x_{t-1},t') = \frac{\sreal(x_{t'}) - \sfake(x_{t'})}{\text{sign}(x_{t-1} - \mu_\theta(x_t))}\frac{CS}{||\mu_\text{real}(x_{t'},t')-\hat x_{0|t}||_1},
\end{equation}
where $\text{sign}(x) = 1$ if $x > 0$, and $-1$ otherwise. $C$ and $S$ is the number of channels and spatial locations, respectively. After applying GN towards $\Rdm$, we multiply the scaler $w_{\text{dm}}$ with $\Adm$ to maintain the original amplitude:
\begin{equation}
    w_{\text{dm},t} = \frac{1}{|x_{t-1}-\mu_\theta(x_t)|+\epsilon}\frac{\sigma_{t-1}^2}{\alpha_{t-1}}.
\end{equation}
This formulation ensures a stable optimization signal by reducing the variance inherent in the raw sampling residuals.

When incorporating with other rewards, we found multiply adaptive weight $\betadmt$ based on $w_{\text{dm},t}$ can effectively improve the target score without collapsing by keeping their amplitudes consistent: 
\begin{equation}\label{eq:betadmt}
    \beta_{\text{dm},t} = \frac{||w_{\text{dm},t}||_1}{CS}.
\end{equation}
As $w_{\text{dm},t}$ is pixel-wise, we keep it consistent with the other rewards' dimensions by averaging them out in the sample dimension. As discussed in \Cref{sec:abl_beta}, $\betadmt$ shows significance on both DMDR and GNDMR.  Finally, we rewrite \Cref{eq:Asum} with designed weight as
\begin{equation}\label{eq:Asum_beta}
    A_{\text{sum},t}^i = w_{\text{dm},t}A_{\text{dm},t}^{i,t'} + \beta_{\text{dm},t}\sum_j w_jA_{\text{o}_j,t}^i.
\end{equation}
\Cref{alg:1} outlines the final training procedure. Additional details are provided in the supplementary materials.

\begin{algorithm}[t!]\small
    \caption{GNDMR training procedure}\label{alg:1}
    \DontPrintSemicolon

    \Input{Pretrained real diffusion model $\mu_{\text{real}}$, number of prompts $N$, number of generated samples per prompt $M$ (group size), number inference timesteps $T$.}
    \Output{Distilled generator $G_\theta$.}
    
    \BlankLine
    \tcp*[h]{Initialize generator and fake score estimators from pretrained model}
    
    $G_\theta \leftarrow \text{copyWeights}(\mu_{\text{real}})$, $\mu_{\text{fake}} \leftarrow \text{copyWeights}(\mu_{\text{real}})$\;
    
    \While{train}{
        \tcp*[h]{Sample trajectories}
        
        Sample a batch of prompts $\mathcal{Q} = \{q_1, q_2, \dots, q_N\}$\;
        \For{each $q \in \mathcal{Q}$}{
            Sample a group of M trajectories $\{(x_T^i,x_{T-1}^i,...,x_0^i)\}_{i=1}^M$
        }

        \tcp*[h]{Update fake score estimation model}\;
        Sample generated timesteps $t$ and diffused timesteps $t'$\;
        $x_{t'}$ = forwardDiffusion(stopgrad($G_\theta(x_t)$),$t'$)\;
        $\mathcal{L}_{\text{denoise}}=$ denoisingLoss($\mu_\text{fake}(x_{t'},t')$, stopgrad($G_\theta(x_t)$)) \tcp*[h]{\Cref{eq:denoise}}\;
        $\mu_\text{fake}=$ update($\mu_\text{fake}$, $\mathcal{L}_{\text{denoise}}$)\;
        \BlankLine
        \tcp*[h]{Compute rewards and advantages}\;
        \For{each $q \in \mathcal{Q}$}{
            Sample generated timesteps $t$ and diffused timesteps $t'$\;
            \For{$i = 1$ \KwTo $M$}{
                $\Rdm^i=\Rdm(x_t^i,x_{t-1}^i,t')$, $R^i_{o} = RM(x_0^i)$ \tcp*[h]{\Cref{eq:Rdm_practice}} \;
                $A_{\text{dm},t}^{i,t'} = \text{GN}(\Rdm^i)$, $A_{\text{o},t}^i = \text{GN}(R^i_{o})$ \tcp*[h]{\Cref{eq:Adm}, \Cref{eq:Ao}}\; 
                $A_{\text{sum},t}^i=$ weightedAdd($A_{\text{dm},t}^{i,t'}$, $A_{\text{o},t}^i$) \tcp*[h]{\Cref{eq:Asum_beta}}\;
            }
        }
        \BlankLine
        \tcp*[h]{Update generator}\;
        \For{train generator loop}{
            Use the same generated timesteps $t$ when compute rewards\;
            $\mathcal{J}_{\text{GNDMR}}(\theta)=$ GRPOLoss($A_\text{sum}$, $r_t(\theta)$) \tcp*[h]{\Cref{eq:loss_GNDMR}}\;
            $G_\theta=$ update($G_\theta$, $\mathcal{J}_{\text{GNDMR}}(\theta)$)\;
        }
    }
\end{algorithm}

\section{Experiments}
\textbf{Experiment Setting.} The distillation is conducted on the LAION-AeS-6.5+\cite{laion} solely with its prompts. We use DFN-CLIP\cite{dfn-clip} and HPSv2.1\cite{hps} as reward models by default, where HPS represents aesthetic quality and CLIP captures image–text alignment, jointly guiding DMD toward more diverse and semantically aligned modes. Meanwhile, to reduce sampling overhead, we propose a variant termed GNDMR-IS, which updates the generator twice for each sampling by importance sampling estimator. The trained distilled models are all flow-based\cite{fm,rectified-flow} and support inference on stochastic sampling\cite{lcm}. We also consider cold start strategy for faster and stable convergence\cite{dmdr}, more experiment details can be found in the supplementary materials.

\noindent\textbf{Evaluation Metrics.} To comprehensively evaluate our approach, we adopt a diverse set of metrics. Fine-grained image-text semantic alignment is measured via the Human Preference Score (HPS) v2.1 \cite{hps}, while PickScore (PS) \cite{pick-score} gauges overall aesthetic quality and perceptual appeal. To capture broader nuances of human judgment such as object accuracy, spatial relations, and attribute binding, we incorporate the recently proposed Multi-dimensional Preference Score (MPS) \cite{mps}. Beyond perceptual assessments, CLIP Score (CS) \cite{dfn-clip} and FID \cite{fid} serve to quantitatively verify distillation effectiveness. We also compute FID-SD \cite{pcm} by comparing the outputs of all baselines against images generated by the original pre-trained diffusion models.

\subsection{Comparison with State-Of-The-Art (SOTA)}
We validate the text-to-image generation performance in both aesthetics alignment and distillation efficiency on 10K prompts from COCO2014\cite{coco} following the 30K split of karpathy. We compare our 4-step generative models GNDMR and its variant GNDMR-IS (update 2 times generator per sampling by importance sampling strategy) against SD3-Medium\cite{sd3} and SD3.5-Medium\cite{sd35}, as well as other open-sourced SOTA distillation models. We reproduced DMD2\cite{dmd2}, as it serves as the foundational distribution-based model in this domain. Additionally, we implemented DMDR (w/ GRPO)\cite{dmdr}, a pioneer method within the same category. To further evaluate our model’s performance against RL techniques directly applied to the teacher model, we extended the application of Flow-GRPO\cite{flow-grpo} to the base model, positioning this as the ceiling of RL performance in this context.

\noindent\textbf{Quantitative Comparison.} As shown in \Cref{tab:compare_sota}, our quantitative analysis highlights the superiority of GNDMR across three key dimensions. Regarding \textbf{Aesthetics Alignment}, on both SD3 and SD3.5, GNDMR consistently outperforms existing 4-step models. It surpasses the teacher and achieves comparable aesthetic performance to this Flow-GRPO optimized model, underscoring the efficacy of our alignment formulation. Furthermore, it enables \textbf{Highly Efficient, Data-Free Distillation}. Operating without external image datasets, GNDMR achieves a strictly lower FID than the data-free DMDR baseline on both SD3 and SD3.5. For SD3, its remarkably low FID-SD indicates accurate teacher distribution matching without the aesthetic degradation seen in Hyper-SD, striking a superior balance between visual quality and distribution fidelity. Finally, our framework delivers \textbf{Reduced Training Cost}, the GNDMR-IS variant halves training expenses while maintaining highly competitive alignment, demonstrating the accelerated convergence and resource efficiency of our paradigm.

\begin{figure}[h!]
    \centering
        \includegraphics[width=\textwidth]{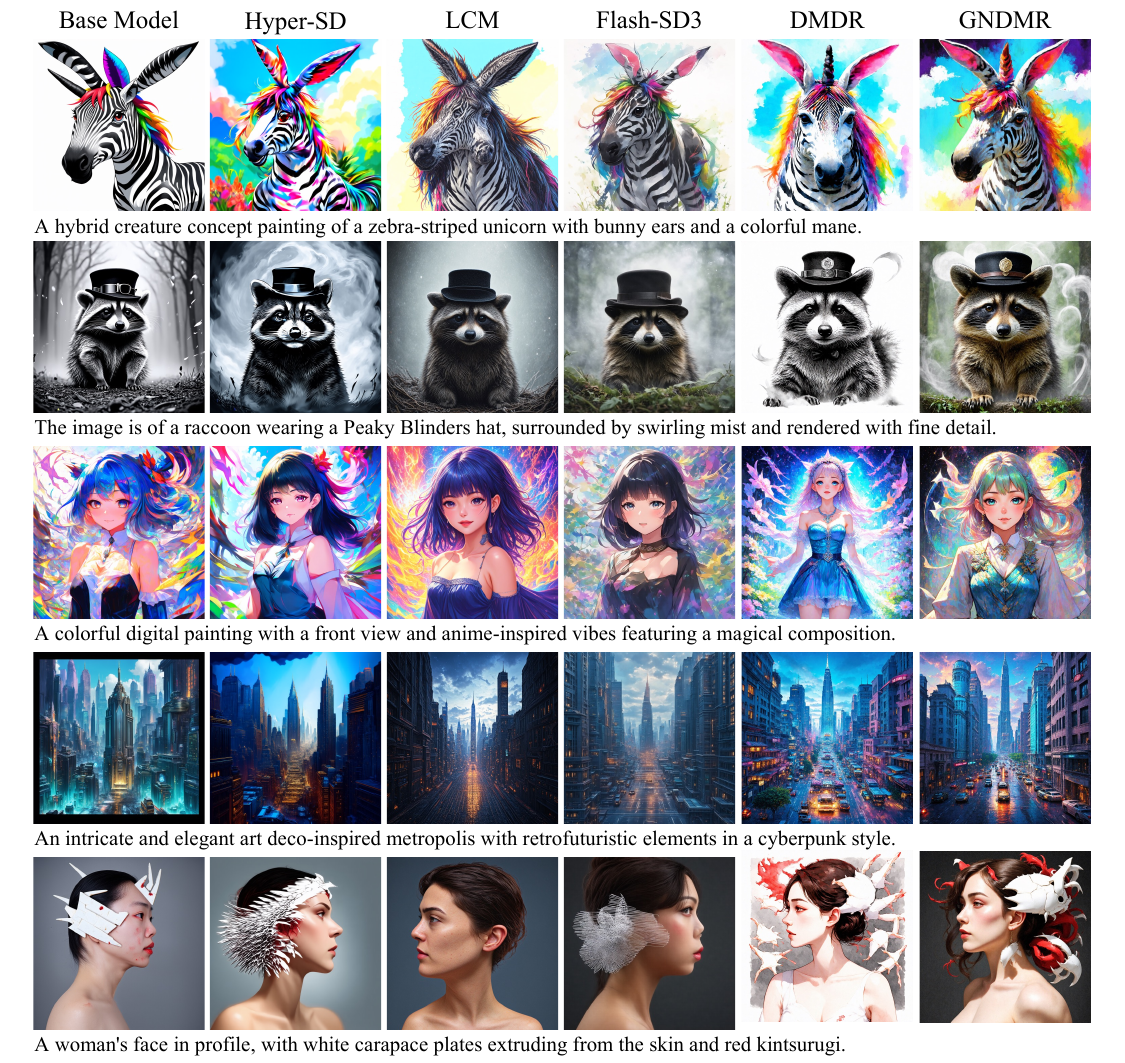}
        \caption{Qualitative Results. Our GNDMR has better aesthetics than other models and fewer artifacts than DMDR.}
        \label{fig:baselines}
\end{figure}

\noindent\textbf{Qualitative Comparison.} As shown in \Cref{fig:baselines}, GNDMR consistently achieves superior aesthetic quality, exhibiting richer details, enhanced color vibrancy, and stronger prompt alignment across diverse styles. Furthermore, our method significantly mitigates the visual artifacts prevalent in DMDR outputs. Our GNDMR demonstrates an optimal balance between high aesthetic appeal and clean image synthesis.

\begin{table*}[!t]
\footnotesize 
\caption{Comparison against state-of-the-art methods. * denotes our reproduced results. \textbf{Img-Free} represents whether the training requires external image data. \textbf{Cost} refers to the product of sample size and sampling iterations. The best results are marked in \colorbox{red!20}{red}, second-best in \colorbox{orange!20}{orange}.}
\label{tab:compare_sota}
\resizebox{\textwidth}{!}{
\begin{tabular}{lcccccccccc} 
\toprule
\textbf{Method} & \textbf{NFE} & \textbf{Res.} & \textbf{Img-Free} & \textbf{HPS}$\uparrow$ & \textbf{PS} $\uparrow$ & \textbf{MPS}$\uparrow$ & \textbf{CS}$\uparrow$ & \textbf{FID}$\downarrow$ & \textbf{FID-SD}$\downarrow$ & \textbf{Cost}$\downarrow$\\
\midrule
\multicolumn{11}{c}{\textbf{Stable Diffusion 3 Medium Comparison}} \\ \midrule
\rowcolor{gray!20}
Base Model (CFG=7) & 50 & 1024 & - & 29.00 & 22.72 & 12.10 & 38.86 & 24.48 & - & - \\
\rowcolor{gray!20}
Flow-GRPO\cite{flow-grpo}* (CFG=7) & 50 & 1024 &-& 30.35 & 22.90 & 12.56 & 38.52 & 26.63 & 12.39 & - \\
Hyper-SD\cite{hyper-sd} (CFG=5) & 8 & 1024 & \xmark & 27.20 & 21.90 & 11.22 & 37.76 & \cellcolor{orange!20}26.94 & \cellcolor{red!20}10.09 & - \\
LCM\cite{lcm} & 4 & 1024 & \xmark & 27.76 & 22.31 & 11.61 & 36.97 & 27.71 & 15.90 & - \\
DMD2\cite{dmd2}* & 4 & 1024 & \xmark & 26.64 & 22.36 & 11.37 & 38.00 & 27.28 & 16.51 & - \\
Flash-SD3\cite{flash-sd3} & 4 & 1024 & \xmark & 27.47 & 22.65 & 11.98 & 38.07 & \cellcolor{red!20}26.01 & \cellcolor{orange!20}12.21 & - \\
DMDR\cite{dmdr}* & 4 & 1024 & \cmark & 29.50 & 22.77 & 11.98 & 38.10 & 29.10 & 14.11 & - \\ \midrule
GNDMR-IS & 4 & 1024 & \cmark & \cellcolor{orange!20}30.00 & \cellcolor{red!20}22.89 & \cellcolor{orange!20}12.48 & \cellcolor{orange!20}38.15 & 28.84 & 12.47 & \cellcolor{red!20}128*4k \\
GNDMR & 4 & 1024 & \cmark & \cellcolor{red!20}\underline{30.37} & \cellcolor{orange!20}\underline{22.88} & \cellcolor{red!20}\underline{12.53} & \cellcolor{red!20}\underline{38.20} & \underline{28.02} & \cellcolor{orange!20}\underline{12.21} & \underline{128*8k} \\
\midrule
\multicolumn{11}{c}{\textbf{Stable Diffusion 3.5 Medium Comparison}} \\ \midrule
\rowcolor{gray!20}
Base Model (CFG=3.5) & 50 & 512 & - & 27.78 & 22.59 & 11.91 & 38.46 & 20.69 & - & - \\
\rowcolor{gray!20}
Flow-GRPO\cite{flow-grpo}* (CFG=3.5) & 50 & 512 & - & 31.81 & 23.24 & 12.93 & 39.21 & 29.27 & 9.39 & - \\
DMD2\cite{dmd2}* & 4 & 512 & \xmark & 30.44 & 22.92 & 12.73 & \cellcolor{red!20}38.59 & 26.64 & \cellcolor{orange!20}14.63 & - \\
DMDR\cite{dmdr}* & 4 & 512 & \cmark & 30.83 & \cellcolor{orange!20}23.07 & 12.80 & 38.22 & 26.05 & 16.73 & - \\ \midrule
GNDMR-IS & 4 & 512 & \cmark & \cellcolor{orange!20}30.88 & 22.94 & \cellcolor{orange!20}12.86 & \cellcolor{orange!20}38.39 & \cellcolor{orange!20}25.60 & 16.68 & \cellcolor{red!20}128*3k \\
GNDMR & 4 & 512 & \cmark & \cellcolor{red!20}\underline{31.25} & \cellcolor{red!20}\underline{23.15} & \cellcolor{red!20}\underline{12.93} & \cellcolor{red!20}\underline{38.59} & \cellcolor{red!20}\underline{24.44} & \cellcolor{red!20}\underline{13.93} & \underline{128*6k} \\
\bottomrule
\end{tabular}}
\end{table*}

\subsection{Importance Sampling Correction}\label{sec:IS}
One significant advantage of treating distribution matching as a reward is the ability to leverage the Importance Sampling (IS) estimator \cite{approximately}. This enables multi-step updates for the student model from a single sampling iteration, effectively addressing the high sample demand typically associated with Reinforcement Learning (RL). Our experiments, conducted on SD3-Medium (512$\times$512), employ HPSv2.1 as an auxiliary reward and report HPSv2.1 on HPDv2 test prompts\cite{hps}.

\begin{figure}[t!]
    \centering
    \begin{subfigure}[b]{0.48\linewidth}
        \centering
        \includegraphics[width=\textwidth]{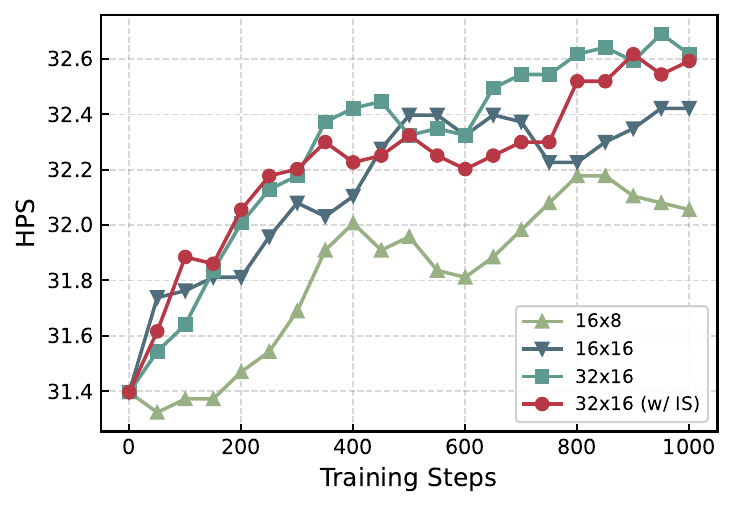}
        \caption{Equal Updates Comparison}
        \label{fig:IS_steps}
    \end{subfigure}
    \hfill 
    \begin{subfigure}[b]{0.48\linewidth}
        \centering
        \includegraphics[width=\textwidth]{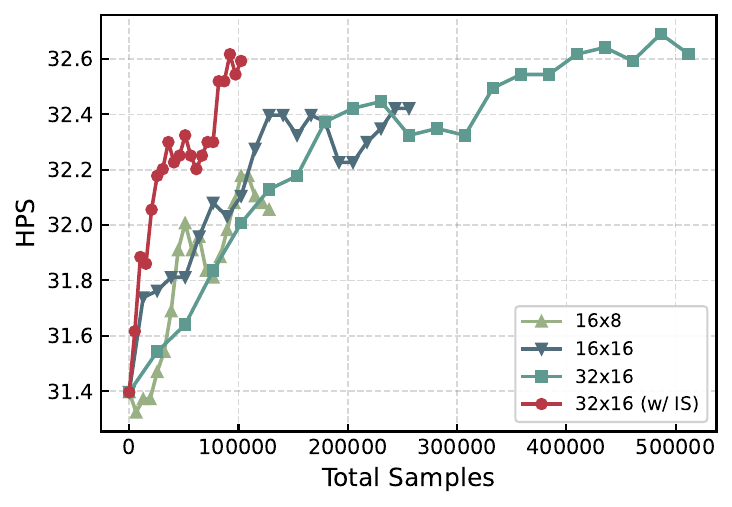}
        \caption{Equal Sampling Budget Comparison}
        \label{fig:IS_samples}
    \end{subfigure}
    \caption{Importance Sampling (IS) improves sampling efficiency. (a) With the same number of training steps, larger batch sizes lead to better reward optimization, but (b) they also require more samples. By introducing IS, the 32×16 (w/ IS) setting achieves comparable performance under a reduced sampling budget.}
\end{figure}

As illustrated in \Cref{fig:IS_steps}, while increasing the batch size (e.g., from $16 \times 8$ to $32 \times 16$) enhances reward optimization per training step, it substantially raises the sampling cost. By implementing 5 training iterations per sample with a clip range of $\eta=0.5$, our 32$\times$16 (w/ IS) configuration achieves a convergence rate and final performance comparable to the standard 32$\times$16 setup, while requiring significantly fewer total samples even less than the baseline 16$\times$8 configuration (see \Cref{fig:IS_samples}).

\begin{table}[h]
  \centering
  \begin{minipage}[t]{0.56\textwidth}
    \centering
    \caption{Ablation on Group Normalization (GN) on $\Rdm$. We first only distill model for 500 iteration then continue training with different rewards.}
    \label{tbl:norm_Rdm}
    \small 
    \begin{tabular}{c c | cc | cc}
      \toprule
      \multirow{2}{*}{\textbf{Method}} & \textbf{only $\Rdm$} & \multicolumn{2}{c|}{\textbf{+HPS}} & \multicolumn{2}{c}{\textbf{+PS}} \\
       & \textbf{FID}$\downarrow$ & \textbf{FID}$\downarrow$ & \textbf{HPS}$\uparrow$ & \textbf{FID}$\downarrow$ & \textbf{PS}$\uparrow$ \\
      \midrule
      GNDMR & 23.07 & 24.47 & 30.22 & 22.32 & 22.76 \\
      w/o GN & 24.94 & 25.40 & 30.39 & 24.61 & 22.71 \\
      \bottomrule
    \end{tabular}
  \end{minipage}
  \hfill 
  \begin{minipage}[t]{0.42\textwidth}
    \centering
    \caption{Ablation on sampling strategy (left) and timestep intervals of group normalization (right).}
    \label{tbl:sample_strategy}
    \small
    \vspace{4.5pt}
    \begin{tabular}[t]{ccc}
        \toprule
        \textbf{$t'$} & \textbf{$t$}  & \textbf{FID} $\downarrow$ \\
        \midrule
             &      & 24.94 \\
        \oldcmark &      & 24.34 \\
        \oldcmark & \oldcmark & 23.07  \\
        \bottomrule
    \end{tabular}
     \hfill 
    \begin{tabular}[t]{lc}
        \toprule
        \textbf{Interval} & \textbf{FID} $\downarrow$ \\
        \midrule
        $[0,300]$ & 23.71 \\
        $[300,600]$ & 23.46 \\
        $[600,1000]$ & 23.43 \\
        \bottomrule
    \end{tabular}
  \end{minipage}
\end{table}

\subsection{Ablation Study}
We explore the effectiveness of $\Rdm$ from multiple perspectives through ablation studies below. By default, we use SD3-Medium with $512 \times 512$ and first perform 500 iterations vanilla DMD for fast training and observation, the results were reported on COCO30K. 

\noindent\textbf{Effect of group normalization on $\Rdm$.}
The primary advantage of our GNDMR lies in applying Group Normalization (GN) to $\Rdm$. To investigate this effect, we conduct the experiments shown in \Cref{tbl:norm_Rdm}. During the first 500 distillation training iterations, GNDMR results in a lower FID, providing a better initialization for subsequent optimization. In the follow-up training, where HPS and PS are separately introduced as additional rewards, GNDMR continues to achieve lower FID while maintaining comparable performance on the target rewards.

\noindent\textbf{Sampling strategy of generation timestep $t$ and diffused timestep $t'$.}\label{sec:abl_timestep} Since Group Normalization (GN) is applied to $\Rdm$, the same $t'$ is shared within each group, as $t'$ determines the noise level of the diffused samples. We further examine whether sharing the same $t$ within a group affects performance. The first row in \Cref{tbl:sample_strategy} (left) corresponds to the vanilla DMD baseline. Sharing both $t'$ and $t$ within a group yields the best performance, consistent with \Cref{eq:Rdm2}, since $\Rdm$ depends on both $t'$ and $t$.

\noindent\textbf{Effect of diffused timestep intervals of group normalization on $\Rdm$.}\label{sec:abl_interval}
As shown in \Cref{tbl:sample_strategy} (right), larger timestep intervals correspond to higher noise levels, which increase the variance of $\Rdm$ estimation as shown in \Cref{fig:std}. Applying GN to regions with larger interval values effectively stabilizes the estimation and leads to lower FID.

\begin{figure}[htbp]
    \centering
    \begin{subfigure}[b]{0.48\linewidth}
        \centering
        \includegraphics[width=\textwidth]{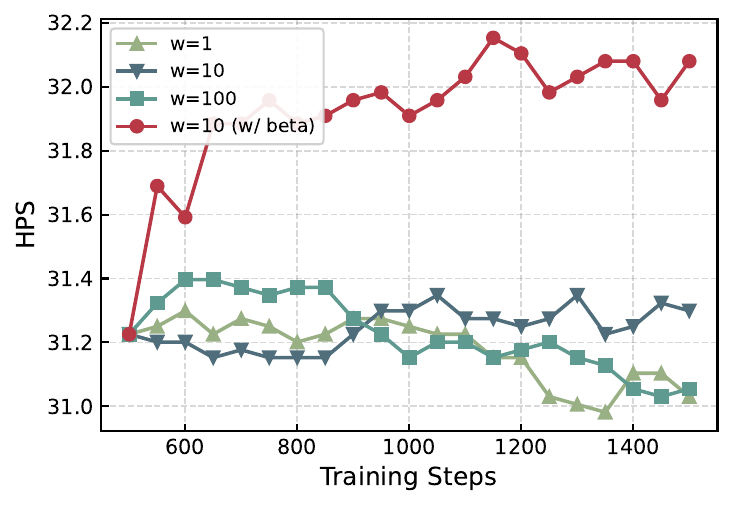}
        \caption{Effect of $\betadmt$ on DMDR}
        \label{fig:dmdr}
    \end{subfigure}
    \hfill 
    \begin{subfigure}[b]{0.48\linewidth}
        \centering
        \includegraphics[width=\textwidth]{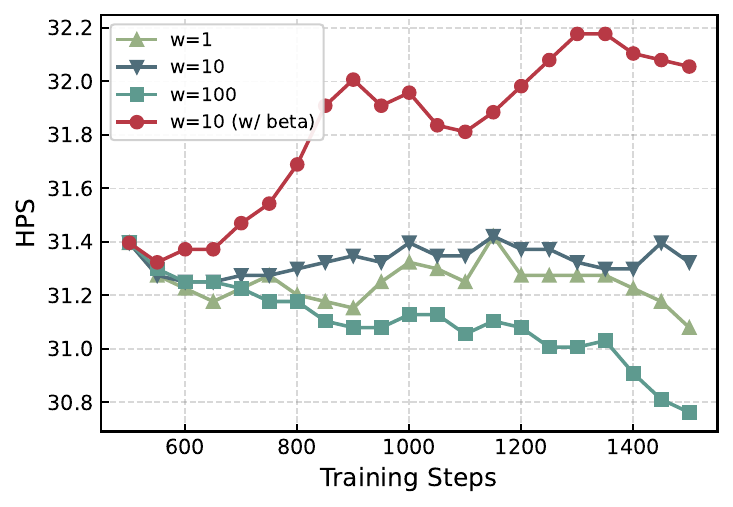}
        \caption{Effect of $\betadmt$ on GNDMR}
        \label{fig:dmrs}
    \end{subfigure}
    \caption{Ablation study on the effect of $\beta_{\text{dm},t}$ and the weighting factor $w$. "w/ beta" denotes the configuration using $\beta_{\text{dm},t}$ as defined in \Cref{eq:betadmt}, whereas other variants set $\beta_{\text{dm},t} = 1$. The HPS is evaluated on the HPDv2 test prompts. The results demonstrate that incorporating $\beta_{\text{dm},t}$ leads to more stable and superior reward optimization compared to static weighting under both DMDR and GNDMR setting.}
    \label{fig:betadm}
\end{figure}

\noindent\textbf{Effect of $\betadmt$.}\label{sec:abl_beta} As illustrated in \Cref{fig:betadm}, it is challenging to consistently improve the target reward through simple static weighting $w$. This difficulty arises because the distillation matching incorporates a dynamic coefficient $w_{\text{dm},t}$ that evolves during training. Without scaling the reward by the same level coefficient $\beta_{\text{dm},t}$, it is nearly impossible to maintain a proper balance between the two terms using a fixed weight $w$, often leading to unstable optimization or suboptimal reward gains. By contrast, applying $\beta_{\text{dm},t}$ to the reward term ensures a synchronized weighting scheme, allowing the reward to improve steadily without collapse.

\section{Conclusion}
In this work, we present a novel framework for improving diffusion distillation by re-conceptualizing distribution matching as a reward. Our approach bridges the gap between diffusion distillation and RL, resulting in more efficient and stable training. Through extensive experiments, we demonstrate that GNDM outperforms vanilla DMD, achieving a notable reduction in FID scores. Furthermore, the GNDMR framework, which integrates additional rewards, achieves an optimal balance between aesthetic quality and fidelity. Our method offers a flexible, efficient, and stable framework for real-time, high-fidelity image synthesis, while also providing a novel direction for the application of the latest RL techniques in diffusion model distillation, paving the way for future advancements in the field.

\bibliographystyle{unsrtnat}
\bibliography{paper}

@String(AAAI  = {AAAI})

@inproceedings{sd3,
  title={Scaling rectified flow transformers for high-resolution image synthesis},
  author={Esser, Patrick and Kulal, Sumith and Blattmann, Andreas and Entezari, Rahim and M{\"u}ller, Jonas and Saini, Harry and Levi, Yam and Lorenz, Dominik and Sauer, Axel and Boesel, Frederic and others},
  booktitle={Forty-first international conference on machine learning},
  year={2024}
}

@inproceedings{DiT,
  title={Scalable diffusion models with transformers},
  author={Peebles, William and Xie, Saining},
  booktitle={Proceedings of the IEEE/CVF international conference on computer vision},
  pages={4195--4205},
  year={2023}
}

@inproceedings{ldm,
  title={High-resolution image synthesis with latent diffusion models},
  author={Rombach, Robin and Blattmann, Andreas and Lorenz, Dominik and Esser, Patrick and Ommer, Bj{\"o}rn},
  booktitle={Proceedings of the IEEE/CVF conference on computer vision and pattern recognition},
  pages={10684--10695},
  year={2022}
}

@article{ddpm,
  title={Denoising diffusion probabilistic models},
  author={Ho, Jonathan and Jain, Ajay and Abbeel, Pieter},
  journal={Advances in neural information processing systems},
  volume={33},
  pages={6840--6851},
  year={2020}
}

@article{score-based,
  title={Score-based generative modeling through stochastic differential equations},
  author={Song, Yang and Sohl-Dickstein, Jascha and Kingma, Diederik P and Kumar, Abhishek and Ermon, Stefano and Poole, Ben},
  journal={arXiv preprint arXiv:2011.13456},
  year={2020}
}

@inproceedings{guided_distill,
  title={On distillation of guided diffusion models},
  author={Meng, Chenlin and Rombach, Robin and Gao, Ruiqi and Kingma, Diederik and Ermon, Stefano and Ho, Jonathan and Salimans, Tim},
  booktitle={Proceedings of the IEEE/CVF conference on computer vision and pattern recognition},
  pages={14297--14306},
  year={2023}
}

@article{progress_distill,
  title={Progressive distillation for fast sampling of diffusion models},
  author={Salimans, Tim and Ho, Jonathan},
  journal={arXiv preprint arXiv:2202.00512},
  year={2022}
}

@inproceedings{improved_cm,
	title        = {Improved Techniques for Training Consistency Models},
	author       = {Yang Song and Prafulla Dhariwal},
	year         = 2024,
	booktitle    = {The Twelfth International Conference on Learning Representations},
	url          = {https://openreview.net/forum?id=WNzy9bRDvG}
}

@inproceedings{dmd,
  title={One-step diffusion with distribution matching distillation},
  author={Yin, Tianwei and Gharbi, Micha{\"e}l and Zhang, Richard and Shechtman, Eli and Durand, Fredo and Freeman, William T and Park, Taesung},
  booktitle={Proceedings of the IEEE/CVF conference on computer vision and pattern recognition},
  pages={6613--6623},
  year={2024}
}

@article{sim,
  title={One-step diffusion distillation through score implicit matching},
  author={Luo, Weijian and Huang, Zemin and Geng, Zhengyang and Kolter, J Zico and Qi, Guo-jun},
  journal={Advances in Neural Information Processing Systems},
  volume={37},
  pages={115377--115408},
  year={2024}
}

@inproceedings{guided_sid,
title={Guided Score identity Distillation for Data-Free One-Step Text-to-Image Generation},
author={Mingyuan Zhou and Zhendong Wang and Huangjie Zheng and Hai Huang},
booktitle={The Thirteenth International Conference on Learning Representations},
year={2025},
url={https://arxiv.org/abs/2406.01561}
}

@article{dmd2,
  title={Improved distribution matching distillation for fast image synthesis},
  author={Yin, Tianwei and Gharbi, Micha{\"e}l and Park, Taesung and Zhang, Richard and Shechtman, Eli and Durand, Fredo and Freeman, Bill},
  journal={Advances in neural information processing systems},
  volume={37},
  pages={47455--47487},
  year={2024}
}

@inproceedings{lasro,
  title={Reward fine-tuning two-step diffusion models via learning differentiable latent-space surrogate reward},
  author={Jia, Zhiwei and Nan, Yuesong and Zhao, Huixi and Liu, Gengdai},
  booktitle={Proceedings of the Computer Vision and Pattern Recognition Conference},
  pages={12912--12922},
  year={2025}
}

@article{dipp,
  title={Diff-instruct++: Training one-step text-to-image generator model to align with human preferences},
  author={Luo, Weijian},
  journal={arXiv preprint arXiv:2410.18881},
  year={2024}
}

@article{dmdr,
  title={Distribution Matching Distillation Meets Reinforcement Learning},
  author={Jiang, Dengyang and Liu, Dongyang and Wang, Zanyi and Wu, Qilong and Li, Liuzhuozheng and Li, Hengzhuang and Jin, Xin and Liu, David and Li, Zhen and Zhang, Bo and others},
  journal={arXiv preprint arXiv:2511.13649},
  year={2025}
}

@article{flow-grpo,
  title={Flow-grpo: Training flow matching models via online rl},
  author={Liu, Jie and Liu, Gongye and Liang, Jiajun and Li, Yangguang and Liu, Jiaheng and Wang, Xintao and Wan, Pengfei and Zhang, Di and Ouyang, Wanli},
  journal={arXiv preprint arXiv:2505.05470},
  year={2025}
}

@inproceedings{
adrpo,
title={Adaptive Divergence Regularized Policy Optimization for Fine-tuning Generative Models},
author={Jiajun Fan and Tong Wei and Chaoran Cheng and Yuxin Chen and Ge Liu},
booktitle={The Thirty-ninth Annual Conference on Neural Information Processing Systems},
year={2025},
url={https://openreview.net/forum?id=aXO0xg0ttW}
}

@misc{gdpo,
      title={GDPO: Group reward-Decoupled Normalization Policy Optimization for Multi-reward RL Optimization}, 
      author={Shih-Yang Liu and Xin Dong and Ximing Lu and Shizhe Diao and Peter Belcak and Mingjie Liu and Min-Hung Chen and Hongxu Yin and Yu-Chiang Frank Wang and Kwang-Ting Cheng and Yejin Choi and Jan Kautz and Pavlo Molchanov},
      year={2026},
      eprint={2601.05242},
      archivePrefix={arXiv},
      primaryClass={cs.CL},
      url={https://arxiv.org/abs/2601.05242}, 
}

@article{flash-dmd,
  title={Flash-DMD: Towards High-Fidelity Few-Step Image Generation with Efficient Distillation and Joint Reinforcement Learning},
  author={Chen, Guanjie and Huang, Shirui and Liu, Kai and Zhu, Jianchen and Qu, Xiaoye and Chen, Peng and Cheng, Yu and Sun, Yifu},
  journal={arXiv preprint arXiv:2511.20549},
  year={2025}
}

@inproceedings{decoupled-dmd,
title={Decoupled {DMD}: {CFG} Augmentation as the Spear, Distribution Matching as the Shield},
author={Dongyang Liu and Peng Gao and David Liu and Ruoyi Du and Zhen Li and Qilong Wu and Xin Jin and Sihan Cao and Shifeng Zhang and Steven HOI and Hongsheng Li},
booktitle={The Fourteenth International Conference on Learning Representations},
year={2026},
url={https://openreview.net/forum?id=jBztvOiCKE}
}

@article{refl,
  title={Imagereward: Learning and evaluating human preferences for text-to-image generation},
  author={Xu, Jiazheng and Liu, Xiao and Wu, Yuchen and Tong, Yuxuan and Li, Qinkai and Ding, Ming and Tang, Jie and Dong, Yuxiao},
  journal={Advances in Neural Information Processing Systems},
  volume={36},
  pages={15903--15935},
  year={2023}
}

@inproceedings{dpo,
  title={Diffusion model alignment using direct preference optimization},
  author={Wallace, Bram and Dang, Meihua and Rafailov, Rafael and Zhou, Linqi and Lou, Aaron and Purushwalkam, Senthil and Ermon, Stefano and Xiong, Caiming and Joty, Shafiq and Naik, Nikhil},
  booktitle={Proceedings of the IEEE/CVF Conference on Computer Vision and Pattern Recognition},
  pages={8228--8238},
  year={2024}
}

@article{ddpo,
  title={Training diffusion models with reinforcement learning},
  author={Black, Kevin and Janner, Michael and Du, Yilun and Kostrikov, Ilya and Levine, Sergey},
  journal={arXiv preprint arXiv:2305.13301},
  year={2023}
}

@inproceedings{
pso,
title={Tuning Timestep-Distilled Diffusion Model Using Pairwise Sample Optimization},
author={Zichen Miao and Zhengyuan Yang and Kevin Lin and Ze Wang and Zicheng Liu and Lijuan Wang and Qiang Qiu},
booktitle={The Thirteenth International Conference on Learning Representations},
year={2025},
url={https://openreview.net/forum?id=fXnE4gB64o}
}

@article{hyper-sd,
  title={Hyper-sd: Trajectory segmented consistency model for efficient image synthesis},
  author={Ren, Yuxi and Xia, Xin and Lu, Yanzuo and Zhang, Jiacheng and Wu, Jie and Xie, Pan and Wang, Xing and Xiao, Xuefeng},
  journal={Advances in neural information processing systems},
  volume={37},
  pages={117340--117362},
  year={2024}
}

@inproceedings{tdm,
  title={Learning few-step diffusion models by trajectory distribution matching},
  author={Luo, Yihong and Hu, Tianyang and Sun, Jiacheng and Cai, Yujun and Tang, Jing},
  booktitle={Proceedings of the IEEE/CVF International Conference on Computer Vision},
  pages={17719--17728},
  year={2025}
}

@article{reinforcement,
  title={Simple statistical gradient-following algorithms for connectionist reinforcement learning},
  author={Williams, Ronald J},
  journal={Machine learning},
  volume={8},
  number={3},
  pages={229--256},
  year={1992},
  publisher={Springer}
}

@article{monte,
  title={Monte carlo gradient estimation in machine learning},
  author={Mohamed, Shakir and Rosca, Mihaela and Figurnov, Michael and Mnih, Andriy},
  journal={Journal of Machine Learning Research},
  volume={21},
  number={132},
  pages={1--62},
  year={2020}
}

@inproceedings{approximately,
  title={Approximately optimal approximate reinforcement learning},
  author={Kakade, Sham and Langford, John},
  booktitle={Proceedings of the nineteenth international conference on machine learning},
  pages={267--274},
  year={2002}
}

@article{ppo,
  title={Proximal policy optimization algorithms},
  author={Schulman, John and Wolski, Filip and Dhariwal, Prafulla and Radford, Alec and Klimov, Oleg},
  journal={arXiv preprint arXiv:1707.06347},
  year={2017}
}

@article{hps,
  title={Human preference score v2: A solid benchmark for evaluating human preferences of text-to-image synthesis},
  author={Wu, Xiaoshi and Hao, Yiming and Sun, Keqiang and Chen, Yixiong and Zhu, Feng and Zhao, Rui and Li, Hongsheng},
  journal={arXiv preprint arXiv:2306.09341},
  year={2023}
}

@inproceedings{clipscore,
  title={Clipscore: A reference-free evaluation metric for image captioning},
  author={Hessel, Jack and Holtzman, Ari and Forbes, Maxwell and Le Bras, Ronan and Choi, Yejin},
  booktitle={Proceedings of the 2021 conference on empirical methods in natural language processing},
  pages={7514--7528},
  year={2021}
}

@article{pref-grpo,
  title={Pref-GRPO: Pairwise Preference Reward-based GRPO for Stable Text-to-Image Reinforcement Learning},
  author={Wang, Yibin and Li, Zhimin and Zang, Yuhang and Zhou, Yujie and Bu, Jiazi and Wang, Chunyu and Lu, Qinglin and Jin, Cheng and Wang, Jiaqi},
  journal={arXiv preprint arXiv:2508.20751},
  year={2025}
}

@article{laion,
  title={Laion-5b: An open large-scale dataset for training next generation image-text models},
  author={Schuhmann, Christoph and Beaumont, Romain and Vencu, Richard and Gordon, Cade and Wightman, Ross and Cherti, Mehdi and Coombes, Theo and Katta, Aarush and Mullis, Clayton and Wortsman, Mitchell and others},
  journal={Advances in neural information processing systems},
  volume={35},
  pages={25278--25294},
  year={2022}
}

@article{dfn-clip,
  title={Data filtering networks},
  author={Fang, Alex and Jose, Albin Madappally and Jain, Amit and Schmidt, Ludwig and Toshev, Alexander and Shankar, Vaishaal},
  journal={arXiv preprint arXiv:2309.17425},
  year={2023}
}

@article{lcm,
  title={Latent consistency models: Synthesizing high-resolution images with few-step inference},
  author={Luo, Simian and Tan, Yiqin and Huang, Longbo and Li, Jian and Zhao, Hang},
  journal={arXiv preprint arXiv:2310.04378},
  year={2023}
}

@article{fm,
  title={Flow matching for generative modeling},
  author={Lipman, Yaron and Chen, Ricky TQ and Ben-Hamu, Heli and Nickel, Maximilian and Le, Matt},
  journal={arXiv preprint arXiv:2210.02747},
  year={2022}
}

@article{rectified-flow,
  title={Flow straight and fast: Learning to generate and transfer data with rectified flow},
  author={Liu, Xingchao and Gong, Chengyue and Liu, Qiang},
  journal={arXiv preprint arXiv:2209.03003},
  year={2022}
}

@article{pick-score,
  title={Pick-a-pic: An open dataset of user preferences for text-to-image generation},
  author={Kirstain, Yuval and Polyak, Adam and Singer, Uriel and Matiana, Shahbuland and Penna, Joe and Levy, Omer},
  journal={Advances in neural information processing systems},
  volume={36},
  pages={36652--36663},
  year={2023}
}

@inproceedings{mps,
  title={Learning multi-dimensional human preference for text-to-image generation},
  author={Zhang, Sixian and Wang, Bohan and Wu, Junqiang and Li, Yan and Gao, Tingting and Zhang, Di and Wang, Zhongyuan},
  booktitle={Proceedings of the IEEE/CVF Conference on Computer Vision and Pattern Recognition},
  pages={8018--8027},
  year={2024}
}

@article{fid,
  title={Gans trained by a two time-scale update rule converge to a local nash equilibrium},
  author={Heusel, Martin and Ramsauer, Hubert and Unterthiner, Thomas and Nessler, Bernhard and Hochreiter, Sepp},
  journal={Advances in neural information processing systems},
  volume={30},
  year={2017}
}

@article{pcm,
  title={Phased consistency models},
  author={Wang, Fu-Yun and Huang, Zhaoyang and Bergman, Alexander and Shen, Dazhong and Gao, Peng and Lingelbach, Michael and Sun, Keqiang and Bian, Weikang and Song, Guanglu and Liu, Yu and others},
  journal={Advances in neural information processing systems},
  volume={37},
  pages={83951--84009},
  year={2024}
}

@inproceedings{coco,
  title={Microsoft coco: Common objects in context},
  author={Lin, Tsung-Yi and Maire, Michael and Belongie, Serge and Hays, James and Perona, Pietro and Ramanan, Deva and Doll{\'a}r, Piotr and Zitnick, C Lawrence},
  booktitle={European conference on computer vision},
  pages={740--755},
  year={2014},
  organization={Springer}
}

@misc{sd35,
  author       = {Stability AI},
  title        = {Sd3.5},
  year         = {2024},
  howpublished = {\url{https://github.com/Stability-AI/sd3.5}},
}

@inproceedings{adm,
  title={Adversarial distribution matching for diffusion distillation towards efficient image and video synthesis},
  author={Lu, Yanzuo and Ren, Yuxi and Xia, Xin and Lin, Shanchuan and Wang, Xing and Xiao, Xuefeng and Ma, Andy J and Xie, Xiaohua and Lai, Jian-Huang},
  booktitle={Proceedings of the IEEE/CVF International Conference on Computer Vision},
  pages={16818--16829},
  year={2025}
}

@inproceedings{flash-sd3,
  title={Flash diffusion: Accelerating any conditional diffusion model for few steps image generation},
  author={Chadebec, Clement and Tasar, Onur and Benaroche, Eyal and Aubin, Benjamin},
  booktitle={Proceedings of the AAAI Conference on Artificial Intelligence},
  volume={39},
  number={15},
  pages={15686--15695},
  year={2025}
}

@article{dancegrpo,
  title={Dancegrpo: Unleashing grpo on visual generation},
  author={Xue, Zeyue and Wu, Jie and Gao, Yu and Kong, Fangyuan and Zhu, Lingting and Chen, Mengzhao and Liu, Zhiheng and Liu, Wei and Guo, Qiushan and Huang, Weilin and others},
  journal={arXiv preprint arXiv:2505.07818},
  year={2025}
}

@article{mixgrpo,
  title={Mixgrpo: Unlocking flow-based grpo efficiency with mixed ode-sde},
  author={Li, Junzhe and Cui, Yutao and Huang, Tao and Ma, Yinping and Fan, Chun and Yang, Miles and Zhong, Zhao},
  journal={arXiv preprint arXiv:2507.21802},
  year={2025}
}


\newpage
\appendix
\onecolumn
\clearpage        
\onecolumn        
\appendix         

\section{Implementation Details}
\subsection{Experiment Details}
We optimize both the generator and the fake score network using the AdamW optimizer. By default, the momentum parameters $\beta_{1}$ and $\beta_{2}$ are set to $0.9$ and $0.999$, respectively. The fake score is updated once for every single generator update. All experiments are conducted on 8 NVIDIA H800 GPUs.

\noindent\textbf{SD3-Medium  }
We adopt a constant learning rate of $1 \times 10^{-6}$ for both the generator and the fake score. Gradient norm clipping is applied with a threshold of $1.0$. The models are trained at a resolution of $1024 \times 1024$ with a total batch size of $128$, utilizing a group size of $8$ and $16$ groups. Classifier-Free Guidance (CFG) is set to $7.0$. To accelerate convergence, we initially train for 500 iterations with the Human Preference Score (HPS) weight $w_{hps}=5$ and CLIP Score weight $w_{cs}=5$. Subsequently, we continue training for an additional 7.5k iterations, increasing both weights to $10$ ($w_{hps}=10$, $w_{cs}=10$).

\noindent\textbf{SD3.5-Medium  }
We adopt a constant learning rate of $1 \times 10^{-6}$ for both the generator and the fake score, with gradient norm clipping set to $1.0$. Training is performed at a resolution of $512 \times 512$ with a batch size of $128$, a group size of $8$, and $16$ groups. The CFG is set to $3.5$. The training process is completed within 8k iterations using $w_{hps}=10$ and $w_{cs}=10$.

\subsection{Training Algorithm Details}
For a comprehensive understanding, \Cref{agl:2} details the specific implementation for constructing the weighted advantage aggregation used during our training process.
\begin{algorithm}
\caption{weightedAdd}\label{agl:2}
\begin{lstlisting}
# x_curr, x_next, x_0: current, next and final sample from trajectory
# alpha_next, sigma_next: noise level at diffusion process
# G: trained few-step generator
# timestep: diffused timestep added in forward_diffusion
noise = randn_like(x_curr)
pred_x0 = G(x_curr)
mu_next = alpha_next * pred_x0
x_next = mu_next + sigma_next * noise
# diffused x
noise = randn_like(x_curr)
noisy_x = forward_diffusion(pred_x0, noise, timestep)
# denoise using real and fake denoiser
fake_x0 = mu_fake(noisy_x, timestep)
real_x0 = mu_real(noisy_x, timestep)
# calculate Rdm and  weighting_factor introduced by DMD
dm_factor = abs(pred_x0 - real_x0).mean(dim=[1,2,3], keepdim=True)
R_dm = (real_x0 - fake_x0) / sign(x_next - mu_next) / dm_factor
w_dm = (1 / abs(x_next - mu_next) + 1e-7) * (sigma_next **2 / alpha_next)
# calculate other rewards
for j in range(K):
    R_oj = RewardModel(x_0)
beta_dm = w_dm.mean(dim=[1,2,3], keepdim=True)
# calculate advantages and sum them up
A_dm = GroupNorm(R_dm)
A_oj = GroupNorm(R_oj)
A_sum = w_dm * A_dm
for j in range(K):
     A_sum +=  beta_dm * w_j * A_oj
\end{lstlisting}
\end{algorithm}

\begin{table}[ht]
\centering
\caption{Comparison of different noise initialization strategies.}
\label{tab:noise_init}
\begin{tabular}{@{}lc@{}}
\toprule
\textbf{Strategy} & \textbf{FID $\downarrow$} \\ \midrule
Random                           & 24.68          \\
Shared     & 23.07 \\ \bottomrule
\end{tabular}
\end{table}
\section{Additional Experiments}
In this section, we provide further ablation studies to validate the design choices within our framework. By default, all experiments in this section are conducted using SD3-Medium at a resolution of $512 \times 512$. For rapid evaluation, we solely consider the HPS reward and train GNDM for 500 iterations before executing the full GNDMR process.
\subsection{Effect of Noise Initialization}
When applying GNDM, the noise initialization strategy plays a critical role. Existing approaches diverge on this front: while DanceGRPO\cite{dancegrpo} and MixGRPO\cite{mixgrpo} utilize shared initial noise across all candidates within a group, Flow-GRPO\cite{flow-grpo} employs independent random initialization. Our empirical results, summarized in \Cref{tab:noise_init}, demonstrate that shared noise initialization yields superior, more robust distillation performance, as evidenced by a lower FID score.

\subsection{Design of the Adaptive Weight $\betadmt$}
\begin{figure}[th]
    \centering
    \begin{minipage}[b]{0.48\linewidth}
        \centering
        \includegraphics[width=\textwidth]{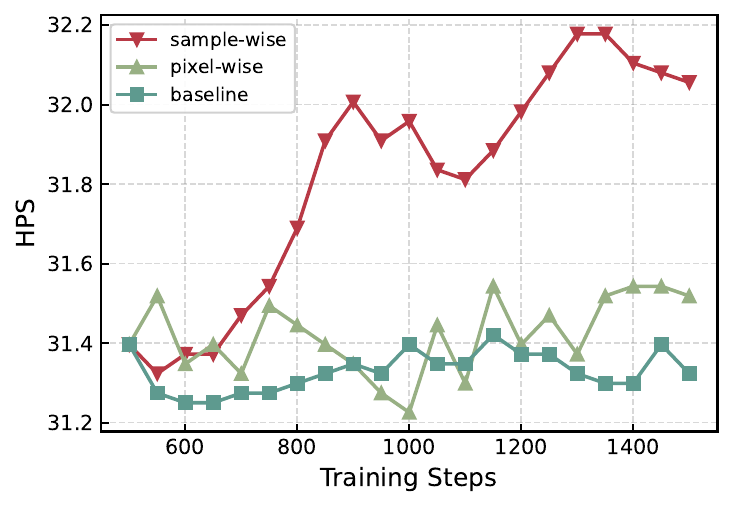}
        \caption{Design of the adaptive weight.}
        \label{fig:design_beta}
    \end{minipage}
    \hfill 
    \begin{minipage}[b]{0.48\linewidth}
        \centering
        \includegraphics[width=\textwidth]{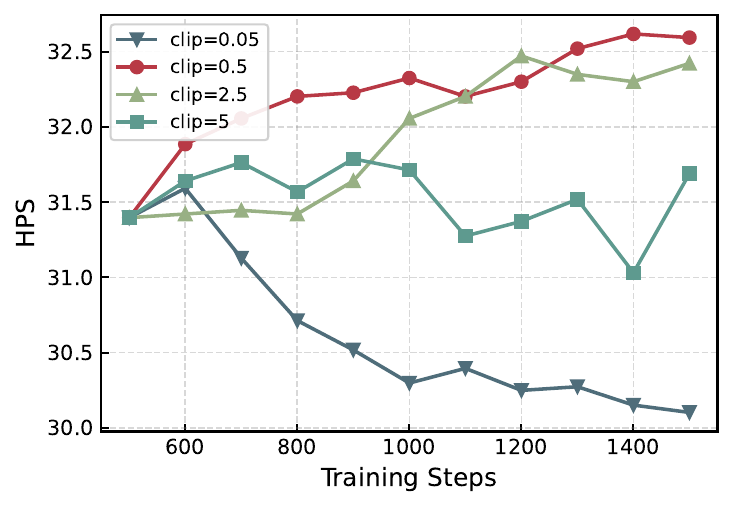}
        \caption{Effect of the clip range.}
        \label{fig:clip_range}
    \end{minipage}
\end{figure}

The coefficient $\beta_{\text{dm},t}$ is designed to calibrate the influence of the distillation weight $w_{\text{dm},t}$. A key technical challenge arises from the dimensionality mismatch: $w_{\text{dm},t}$ is a pixel-wise metric, whereas reinforcement learning rewards are typically sample-wise. To evaluate its necessity and the optimal granularity, we compare three configurations:
\begin{enumerate}
\item A baseline setting with no balancing coefficient (i.e., $\beta_{\text{dm},t} = 1$).
\item A pixel-wise application, where $\beta_{\text{dm},t}$ is directly set to $w_{\text{dm},t}$.
\item A sample-wise variant, where $\beta_{\text{dm},t}$ is computed by taking the mean of $w_{\text{dm},t}$ over all dimensions except for the batch dimension.
\end{enumerate}
Experimental results in \Cref{fig:design_beta} demonstrate that the sample-wise formulation of $\beta_{\text{dm},t}$ is significantly more effective at improving the HPS than both the pixel-wise and baseline configurations, as it provides a much more stable reward signal across the generated samples.

\subsection{Sensitivity Analysis of the Reward Weight $w_\text{hps}$}
\begin{figure}[th]
    \centering
    \begin{subfigure}[b]{0.48\linewidth}
        \centering
        \includegraphics[width=\textwidth]{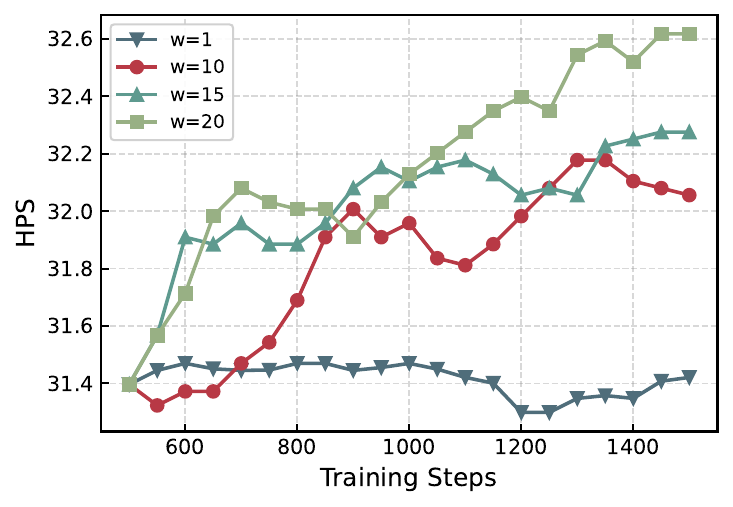}
        \caption{Sensitivity analysis of $w$ with $\betadmt$.}
    \end{subfigure}
    \hfill 
    \begin{subfigure}[b]{0.48\linewidth}
        \centering
        \includegraphics[width=\textwidth]{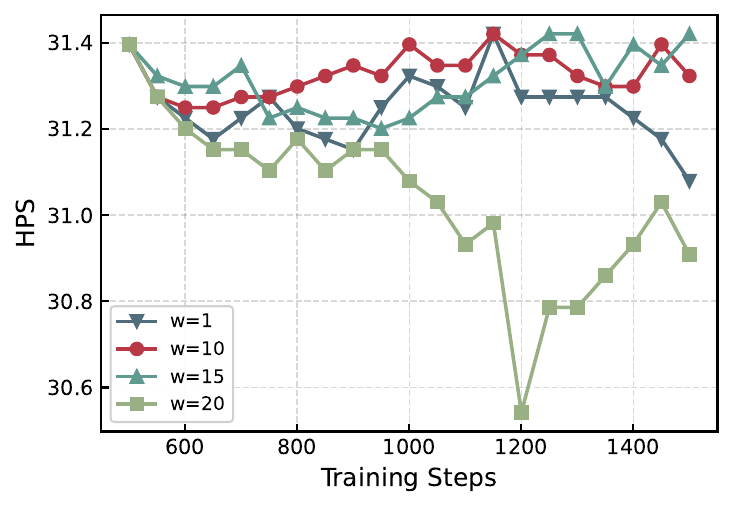}
        \caption{Sensitivity analysis of $w$ without $\betadmt$.}
    \end{subfigure}
    \caption{Sensitivity analysis of the reward weight $w$.}
    \label{fig:sens_beta}
\end{figure}

To further analyze the sensitivity of our adaptive weight, we vary the reward weight for $R_\text{hps}$, denoted as $w_\text{hps}$, across the set $\{1, 10, 15, 20\}$. As shown in \Cref{fig:sens_beta}, enlarging $w_\text{hps}$ improves the overall HPS performance. Furthermore, without integrating $\betadmt$, the model struggles to improve the target HPS consistently, underscoring the need for our adaptive weight $\betadmt$.


\subsection{Sensitivity of Clip Range $\eta$ in Importance Sampling}
The clip range $\eta$ is a pivotal hyperparameter for importance sampling correction. While a larger $\eta$ permits more aggressive policy updates, it often introduces training instability. Conversely, an excessively low $\eta$ restricts learning progress and may fail to correct for distribution shifts from the behavior policy. As shown in \Cref{fig:clip_range}, unlike standard GRPO-based fine-tuning, which typically uses a highly conservative clip range (e.g., $1 \times 10^{-5}$), our distillation framework benefits from a larger $\eta$ to enable rapid, efficient convergence.

\section{Additional Qualitative Results}
\begin{figure}
    \centering
    \includegraphics[width=1\linewidth]{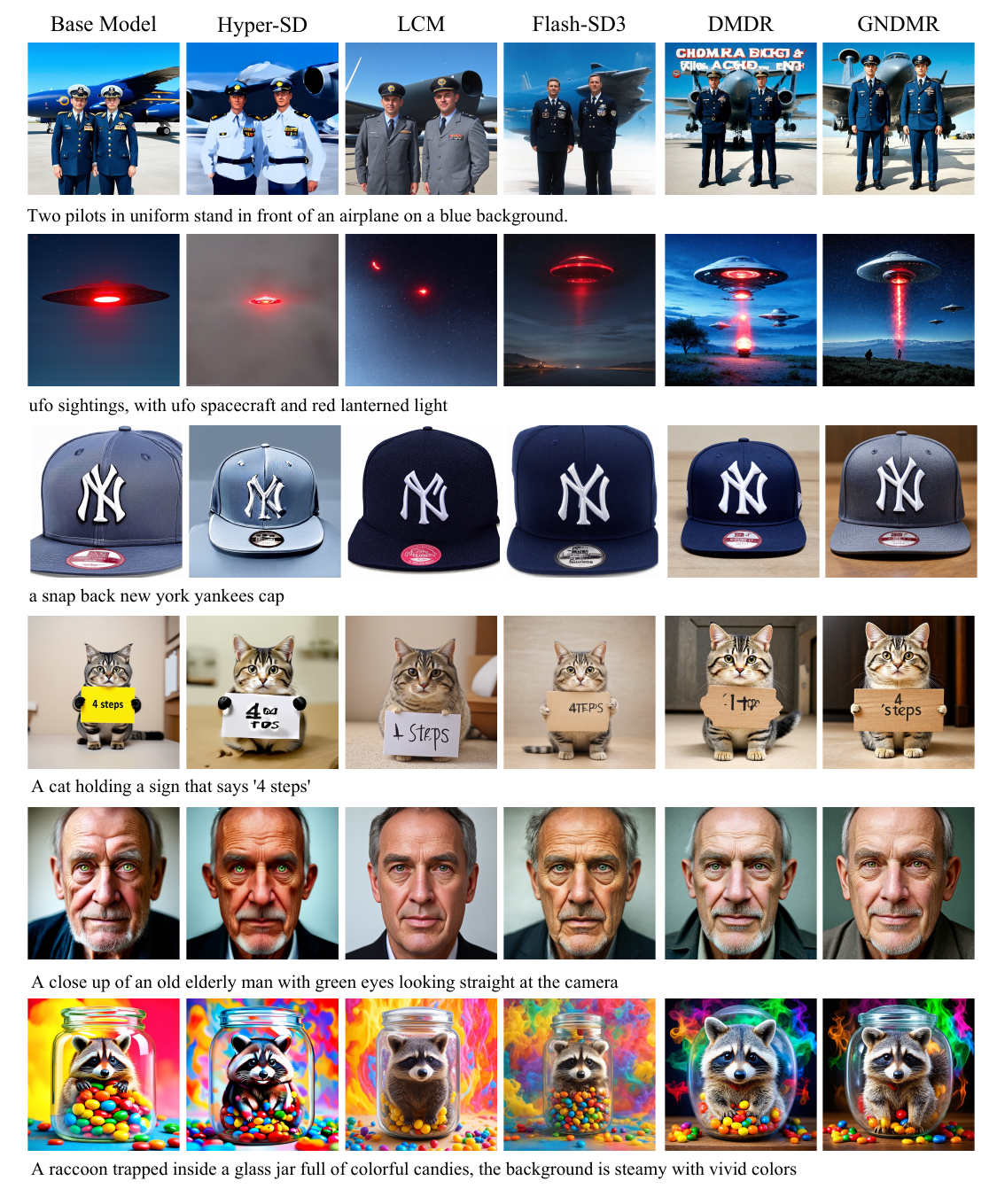}
    \caption{Qualitative results. Text prompts are selected from DMDR \cite{dmdr} (top three rows) and Flash-SD \cite{flash-sd3} (bottom three rows).}
    \label{fig:more_compare}
\end{figure}

\Cref{fig:more_compare} presents additional qualitative comparisons between our proposed GNDMR and several state-of-the-art distillation baselines. Across various complex prompts, GNDMR consistently yields more aesthetically pleasing results with enhanced textural detail and structural integrity.

\end{document}